\documentclass{article}
\usepackage{arxiv}
\usepackage[numbers]{natbib}
\usepackage[utf8]{inputenc}
\usepackage{amsmath, amsfonts, amssymb}
\usepackage[table,xcdraw]{xcolor}
\usepackage{subdef}
\usepackage{algorithm}
\usepackage{algpseudocode}
\usepackage{caption}
\usepackage{subcaption}
\usepackage{multirow}
\usepackage{graphicx}
\usepackage{hyperref}

\title{\textsc{Submodlib}: A Submodular Optimization Library}

\author{Vishal Kaushal \\
Department of Computer Science and Engineering,\\
Indian Institute of Technology Bombay\\
\texttt{vkaushal@cse.iitb.ac.in}\\
\And
Ganesh Ramakrishnan \\
Department of Computer Science and Engineering,\\
Indian Institute of Technology Bombay\\
\texttt{ganesh@cse.iitb.ac.in}\\
\And
Rishabh Iyer \\
  Department of Computer Science\\
  University of Texas at Dallas\\
  \texttt{rishabh.iyer@utdallas.edu} \\
}

\date{}

\begin{document}

\maketitle

\begin{abstract}
    Submodular functions are a special class of set functions which naturally model the notion of representativeness, diversity, coverage etc. and have been shown to be computationally very efficient. A lot of past work has applied submodular optimization to find optimal subsets in various contexts. Some examples include data summarization for efficient human consumption, finding effective smaller subsets of training data to reduce the model development time (training, hyper parameter tuning), finding effective subsets of unlabeled data to reduce the labeling costs, etc. A recent work has also leveraged submodular functions to propose \emph{submodular information measures} which have been found to be very useful in solving the problems of \emph{guided subset selection} and \emph{guided summarization}. In this work, we present \textsc{Submodlib} which is an open-source, easy-to-use, efficient and scalable Python library for submodular optimization with a C++ optimization engine. \textsc{Submodlib} finds its application in summarization, data subset selection, hyper parameter tuning, efficient training and more. Through a rich API, it offers a great deal of flexibility in the way it can be used. Source of \textsc{Submodlib} is available at \href{https://github.com/decile-team/submodlib}{https://github.com/decile-team/submodlib}.
\end{abstract}

\section{Introduction}

Data subset selection or obtaining effective smaller subsets of data finds its use in a variety of applications. Consider deep models for example. While they demonstrate astounding improvements in accuracies on several downstream image, video or text tasks, they pose the following challenges: a) Increased training complexity and computational costs, b) Larger inference time, c) Larger experimental turn around times and difficulty in hyper-parameter tuning, and d) Higher costs and more time for labeling. Some of the ways past work has tried to address one or more of these challenges are through novel network architecture modifications~\citep{srivastava2015training,ba2014deep,hinton2015distilling,iandola2016squeezenet,levi2015age,jaderberg2015spatial}, transfer learning~\citep{donahue2014decaf,pan2010survey,yosinski2014transferable,long2015learning}, zero-shot learning ~\citep{socher2013zero, changpinyo2016synthesized}, one-shot learning~\citep{vinyals2016matching}, activelearning~\citep{settles2010active}, core sets~\citep{agarwal2005geometric} and, in the context of this work, submodular functions~\citep{wei2015submodularity, wei2014submodular, kaushal2019learning}. Yet another application of subset selection is seen in extractive summarization of documents, images or videos wherein a good summary is modeled as an informative, non-redundant and diverse subset of the ground set. Naturally, several past works have leveraged submodular functions and submodular optimization to address document summarization~\citep{lin2012learning}, image collection sumamrization~\citep{tschiatschek2014learning} and video summarization~\citep{kaushal2019demystifying, kaushal2019framework, elhamifar2017online, gygli2015video}.

More recently, ~\cite{kothawade2021prism} have proposed parameterized submodular information measures to address the problem of \emph{guided subset selection} and \emph{guided summarization}. One application of guided subset selection is \emph{targeted learning}, where the goal is to find subsets with rare classes or rare attributes on which the model is under-performing. In practice, there is often a distribution shift between training and testing data. A model's performance on a desired target can be improved (under given additional labeling costs) by augmenting the training data with samples best matching the target distribution from a large pool of unlabeled data. Selecting samples most useful for model training thus leads to efficient cost-effective training of models. Guided data subset selection need not be limited to guiding the subset to be \emph{similar} to a target. There are applications requiring a subset to be different from a set of data points (private set). \emph{Guided summarization}~\citep{kothawade2021prism}, where data ({\em e.g.}, image collection, text, document or video) is summarized for quicker human consumption with specific additional user intent, is one such example. Variants like \emph{privacy-preserving summarization} or \emph{update summarization} require the summary (subset in our case) to be \emph{different} from a private set. 

We release \textsc{Submodlib}, an open-source Python library, which makes it easy for developer or a researcher to use submodular optimization for many tasks including the ones mentioned above. 

In what follows, we first cover some preliminaries around submodular functions and the submodular information measures and then describe \textsc{Submodlib} in details. 

\section{Submodular Functions and Submodular Optimization}
\label{sec:submod}

Given a \emph{ground set} $\Vcal = \{1, 2, 3, \cdots, n\}$ of items (e.g., images, video frames, or sentences) let us define a utility function (set function) $f:2^\Vcal \rightarrow \Re$, which measures how good a subset $X \subseteq \Vcal$ is according to some criteria modeled by the function $f$. In order to find the \emph{best} subset the goal is then to have a subset $\hat{X}$ which maximizes $f$. Below we define two optimization problems relevant to the task of data subset selection and summarization.

\begin{align}
Problem 1: \hat{X} = \max_{X \subseteq \Vcal, s(X) \leq b} f(X)
\end{align}

Problem 1 is knapsack constrained submodular maximization~\citep{sviridenko2004note}. The goal here is to find a subset with a fixed cost. Let $s_1, s_2, \cdots, s_n$ denote the cost of each element in the ground-set. Then $s(X)$, the cost of a subset $X$, is equal to $\sum_{i \in X}s_i(X)$. A special case is cardinality constrained submodular maximization, when the individual costs are $1$~\citep{nemhauser1978analysis}. This a natural model for extracting fixed length summary videos (or a fixed number of keyframes) or to get a subset with a fixed number of data points.

\begin{align}
Problem 2: \hat{X} = \min_{X \subseteq \Vcal, f(X) \geq c} s(X)
\end{align}
This problem is called the \emph{Submodular Cover Problem}~\citep{wolsey1982analysis,iyer2013submodular}. $s(X)$ is the modular cost function, and  $c$ is the coverage constraint. The goal here is to find a minimum cost subset $X$ such that the submodular coverage or representation function covers \emph{information} from the ground set. A special case of this is the set cover problem. Moreover, Problem 2 can be seen as a Dual version of Problem 1~\citep{iyer2013submodular}.

It is easy to see that maximizing a generic set function becomes computationally infeasible as $\Vcal$ grows. However, a special class of set functions, called \textbf{submodular functions} makes this optimization easy. Submodular functions~\citep{fujishige2005submodular} are a special class of set functions $f: 2^\Vcal \rightarrow \Re$. A function $f$ is \emph{submodular} \citep{fujishige2005submodular} if for all $A, B \subseteq \Vcal$, it holds that 

$$f(A) + f(B) \geq f(A \cup B) + f(A \cap B)$$

An identical characterization of submodularity is that they exhibit a "diminishing returns" property: given subsets $A \subseteq B$ and an item $i \notin B$, submodular functions must satisfy $$f(i | A) \triangleq f(A \cup i) - f(A) \geq f(i | B)$$

That is, adding some instance $x$ to the subset $A$ provides more gain in terms of the target function than adding $x$ to a larger subset $B$, where $A \subseteq B$. Informally, since $B$ is a superset of $A$ and already contains more information, adding $x$ will not help as much. This "diminishing returns" property makes submodularity suitable for modeling characteristics such as diversity, coverage, importance and representation. Several diversity and coverage functions are thus submodular, since they satisfy this diminishing returns property. Furthermore, $f$ is supermodular if $-f$ is submodular, and $f$ is said to be {monotone} if $f(X) \leq f(Y)$, $\forall X\subseteq Y\subseteq \Vcal$ (equivalently, $f(j|A) \geq 0$ for all $j \notin A$ and $A \subseteq \Vcal$). 

Submodular functions admit simple and scalable greedy algorithms with constant factor approximation guarantees~\citep{nemhauser1978analysis}. They enable efficient optimization algorithms with guarantees both in the minimization \citep{fujishige2005submodular,iyer2013fast} and maximization settings \citep{krauseSubmodularFunctionMaximization2013, leeNonmonotoneSubmodularMaximization2009, buchbinderTightLinearTime2015}. Using a greedy algorithm to optimize a monotone submodular function (for selecting a subset) gives a lower-bound performance guarantee of a factor of $1 - 1/e$ of optimal ~\citep{nemhauser1978analysis} to Problem 1, and in practice these greedy solutions are often within a factor of 0.98 of the optimal ~\citep{krause2008optimizing}. This makes it advantageous to formulate (or approximate) the objective function for data selection as a submodular function.

Submodularity is a rich yet tractable sub-field of non-linear combinatorial optimization which ensures tractable algorithms and nice connections to convexity and concavity ~\citep{bach2011learning,lovasz1983submodular,iyer2015polyhedral,lovasz1983submodular,iyer2015polyhedral,iyer2020concave}. Submodular functions generalize a number of combinatorial and information theoretic functions such as entropy, set cover, facility location, graph cut, and provide a general class of expressive models. They model aspects like diversity, coverage, information \citep{linClassSubmodularFunctions2011, tschiatschekLearningMixturesSubmodular2014a}, attractive potentials \citep{iyerOptimalAlgorithmsConstrained2014} and cooperation \citep{jegelkaSubmodularitySubmodularEnergies2011}. Due to close connections between submodularity and entropy, submodular functions can also be viewed as \emph{information functions}~\citep{zhang1998characterization}.

For these reasons they have been applied extensively in sensor placement \citep{krauseRobustSubmodularObservation, krauseNearoptimalNonmyopicValue, krauseNearOptimalSensorPlacementsa}, structured learning of graphical models \citep{narasimhanPAClearningBoundedTreewidth2004}, social networks \citep{kempeMaximizingSpreadInfluence2003}, document summarization~\citep{lin2010multi,lin2011class,lin2012learning,li2012multi,chali2017towards,yao2017recent}, image collection summarization~\citep{tschiatschek2014learning}, video summarization~\citep{gygli2015video, zhang2016video,Gygli2015VideoSB,Kaushal2019DemystifyingMV,Kaushal2019AFT,xuGazeenabledEgocentricVideo2015,bairi2015summarization,gygli2015video,kaushal2019framework}, data subset selection and active learning~\citep{wei2015submodularity,liu2015svitchboard,kaushal2019learning} etc. and have been used to achieve state-of-the-art results. Though more recent state-of-the-art summarization techniques use deep learning for modeling \emph{importance} and \emph{relevance}, they are often complemented by submodular functions and Determinantal Point Processes (DPPs)~\citep{kulesza2012determinantal} to represent diversity, representation and coverage~\citep{zhang2016video,cho2019improving,tschiatschek2014learning,vasudevan2017query}.

Below we give an overview of different categories of submodular functions followed by their expressions and characteristics.

\subsection{Functions modeling representation} Representation based functions attempt to directly model representation, in that they try to find a representative subset of items, akin to centroids and medoids in clustering. 

\subsubsection{Facility Location}

The Facility-Location function (FL) ~\citep{mirchandani1990discrete} is closely related to k-medoid clustering. It is defined as  

$$f_{FL}(X) = \sum_{i \in \Vcal} \max_{j \in X} s_{ij}$$

where $i$ is an element from the ground set $\Vcal$ and $s_{ij}$ measures the similarity between element $i$ and element $j$. For each data point $i$ in the ground set $\Vcal$, we compute the representative from subset $X$ which is closest to $i$ and add these similarities for all data points. In a more generic setting, the set whose representation is desired (we call it represented set $\Ucal$) may be different from the set whose subset is desired (i.e. the ground set $\Vcal$). The expression for Facility-Location function in this generic setting then becomes 

$$f_{FL}(X) = \sum_{i \in \Ucal} \max_{j \in X} s_{ij}$$. 

Facility Location is monotone submodular. Note that the Facility Location function requires computing a $O(n^2)$ similarity function. However, as shown in~\cite{wei2014fast}, we can approximate this with a nearest neighbor graph, which will require much less storage, and also can run much faster for large ground set sizes. 

\subsubsection{Graph Cut}

We define the graph-cut family of functions (GC) as 

$$f_{GC}(X) = \sum_{i \in \Vcal, j \in X} s_{ij} - \lambda \sum_{i, j \in X} s_{ij}$$

where $\lambda$ governs the trade-off between representation and diversity. When $\lambda$ becomes large, graph cut function also tries to model diversity in the subset. For $\lambda < 0.5$ it is monotone submodular. For $\lambda > 0.5$ it is non-monotone submodular. Like Facility location, in the more generic setting, the set whose representation is desired (i.e. the represented set $\Ucal$) may be different from the ground set $\Vcal$ whose subset is desired. The expression for Graph Cut function then becomes 

$$f_{GC}(X) = \sum_{i \in \Ucal, j \in X} s_{ij} - \lambda \sum_{i, j \in X} s_{ij}$$

Graph Cut function is similar to the Facility Location and Saturated Coverage in terms of its modeling behaviour.

\subsection{Functions modeling diversity} 

Diversity based functions attempt to obtain a diverse set of keypoints. There is a subtle difference between the notion of diversity and the notion of representativeness. While diversity \emph{only} looks at the elements in the chosen subset, representativeness also worries about their similarity with the remaining elements in the superset. For example, an outlier point will be preferred by a diverse subset but not by a representative subset. 

\subsubsection{Dispersion Functions}

The goal is to have minimum similarity across elements in the chosen subset by maximizing minimum pairwise distance between elements. This is called Minimum Disparity function (DMin). Denote $d_{ij}$ as a distance measure between element $i$ and $j$. Define a set function 

$$f_{DMin}(X) = \min_{i, j \in X} d_{ij}$$

\textbf{This function is not submodular}, but can be still be efficiently optimized via a greedy algorithm~\citep{dasgupta2013summarization}. It is easy to see that maximizing this function involves obtaining a subset with maximal minimum pairwise distance, thereby ensuring a diverse subset of snippets or keyframes. Similar to the Minimum Disparity, we can define two more variants. The first is Disparity Sum (DSum), which can be defined as 

$$f_{DSum}(X) = \sum_{i, j \in X} d_{ij}$$

It models diversity by computing the sum of pairwise distances of all the elements in a subset. This is a supermodular function. The second is, Disparity Min-Sum (DMinSum) which is a combination of the two forms of models. This is defined as 

$$f_{DMinSum}(X) = \sum_{i \in X} \min_{j \in X} d_{ij}$$

Disparity Min-Sum function is submodular~\citep{chakraborty2015adaptive}. 

\subsubsection{Determinantal Point Processes}

A common choice of diversity models used in literature are determinantal point processes (DPP)~\citep{kulesza2012determinantal}, defined as

$$p(X) = \mbox{Det}(L_X)$$ 

where $L$ is a similarity kernel matrix, and $L_X$ denotes a submatrix of $L$ with the rows and columns of $L$ indexed with elements in $X$. It turns out that a close variant, Log Determinant $f(X) = \log p(X)$ is submodular, and hence can be efficiently optimized via the Greedy algorithm. The log-determinant function (LogDet) can thus be defined as 

$$f_{LogDet}(X) = \log\det(L_X)$$

Unlike the Dispersion functions, this requires computing the determinant and is $O(n^3)$ where $n$ is the size of the ground set. This function is not computationally feasible for large scale. 

\subsection{Functions modeling coverage}

This class of functions model notions of coverage, i.e. try to find a subset of the ground set $X$ which covers a set of \emph{concepts}. 

\subsubsection{Set Cover}

For a subset $X$, its Set Cover evaluation (SC) is defined as 

$$f_{SC}(X) = w(\cup_{x \in X} \gamma(x)) = w(\gamma(X))$$ 

where $\gamma(X)$ refers to the set of concepts covered by $X$. Thus the set of all concepts $\mathcal{C} = \gamma(\mathcal{V})$. $w$ is a weight vector in $\Re^{|\mathcal{C}|}$. Intuitively, each element in $\mathcal{V}$ \emph{covers} a set of elements from the concept set $C$ and hence $w(\gamma(X))$ is total weight of concepts covered by elements in $X$. Note that $\gamma(A \cup B) = \gamma(A) \cup \gamma(B)$ and hence $f(A \cup B) = w(\gamma(A \cup B)) = w(\gamma(A) \cup \gamma(B))$. 

Alternatively we can also view the function as follows. Let $\mathcal{C}$ be the set of all concepts (that is, $\mathcal{C} = \gamma(\mathcal{V})$) and $c_u(i)$ denote whether the concept $u \in \mathcal{C}$ is covered by the element $i \in \mathcal{V}$ (that is, $c_u(i) = 1$ if $u \in \gamma(\{i\})$ and is zero otherwise). We then define $c_u(X) = \sum_{x\in X} c_u(x)$ as the count of concept $u$ in set $X$, and the weighted set cover can then be written as 

$$f_{SC}(X) = \sum_{u \in \mathcal{C}} w_u \min(c_u(X), 1)$$

Set Cover function is monotone submodular.

\subsubsection{Probabilistic Set Cover}

Probabilistic Set Cover (PSC) is defined as 

$$f_{PSC}(X) = \sum_{u \in \mathcal{C}} w_u(1 - P_u(X))$$ 

where $\mathcal{C}$ is the set of concepts, $w_u$ is the weight of the concept $u$ and $P_u(X) = \prod_{j \in X} (1 - p_{uj})$ where $p_{xu}$ is the probability with which concept $u$ is covered by element $x$. Thus, $P_u(X)$ is the probability that $X$ \textbf{doesn't} cover concept $u$. In other words, 

$$f_{PSC}(X) = \sum_{u \in \mathcal{C}} w_u(1 - \prod_{x \in X} (1 - p_{xu}))$$

Intuitively, Probabilistic Set Cover function is a softer version of the Set Cover function, which allows for probability of covering concepts, instead of a binary yes/no, as is the case with Set Cover function. Similar to the Set Cover function, this function models the coverage aspect of the candidate summary (subset), viewed stochastically and is also monotone submodular.

\subsubsection{Feature-based Functions}

Feature-based functions (FB) are another class of coverage functions. These are essentially sums of concave over modular functions defined as, 

$$f_{FB}(X) = \sum_{f \in F} w_f g(m_f(X))$$ 

where $g$ is a concave function, ${m_f}$ are a set of feature scores, and $f \in F$ are features. In case of images, features could be, for example, the features extracted from the second last layer of a ConvNet. Examples of $g$ include square-root, log and inverse function. Feature-based functions model the notion of coverage over features.

\section{Submodular Information Measures}
\label{sec:sim}

Next we present the notion of submodular information measures as introduced by ~\cite{kothawade2021prism}.

\subsection{Submodular Conditional gain (CG)} 

Given sets $\Acal, \Pcal \subseteq \Vcal$, the submodular conditional gain (CG), $f(\Acal | \Pcal)$, is the gain in function value by adding $\Acal$ to $\Pcal$. Thus 

$$f(\Acal | \Pcal) = f(\Acal \cup \Pcal) - f(\Pcal)$$

Intuitively, $f(\Acal|\Pcal)$ measures how different $\Acal$ is from $\Pcal$, where $\Pcal$ is the \emph{conditioning set} or the \emph{private set}.

\subsection{Submodular Mutual Information (MI)} 

Given sets $\Acal, \Qcal \subseteq \Vcal$, the submodular mutual information (MI) ~\citep{levin2020online,iyer2021submodular} is defined as 

$$I_f(\Acal; \Qcal) = f(\Acal) + f(\Qcal) - f(\Acal \cup \Qcal)$$

Intuitively, this measures the similarity between $\Qcal$ and $\Acal$ where $\Qcal$ is the query set.

\subsection{Submodular Conditional Mutual Information (CMI)}

Submodular conditional mutual information (CMI) is defined using CG and MI as 

$$I_f(A; Q | P) = f(A | P) + f(Q | P) - f(A \cup Q | P)$$

which is equivalent to 

$$I_f(\Acal; \Qcal | \Pcal) = f(\Acal \cup \Pcal) + f(\Qcal \cup \Pcal) - f(\Acal \cup \Qcal \cup \Pcal) - f(\Pcal)$$

Intuitively, CMI jointly models the mutual similarity between $\Acal$ and $\Qcal$ and their collective dissimilarity from $\Pcal$. 

We now present some instantiations of the above measures which are implemented in \textsc{Submodlib}. We refer to them as \textbullet MI or \textbullet CG or \textbullet CMI where \textbullet\ is the submodular function using which the respective MI, CG or CMI measure is instantiated. 
While different submodular functions naturally model different characteristics such as representation, coverage, {\em etc.}~\citep{Kaushal2019DemystifyingMV,Kaushal2019AFT}, the instantiations presented here additionally model similarity and dissimilarity to query and private sets respectively. These instantiations have parameters $\lambda$, $\eta$ and/or $\nu$, that govern the interplay among different characteristics. In several instantiations, we invoke a similarity matrix $S$ where $S_{ij}$ measures the similarity between elements $i$ and $j$ of sets that will be correspondingly specified. These were first presnted in ~\cite{kothawade2021prism} and we reproduce them below for easy reference.

\subsection{Log Determinant (LogDet) Instantiations} 

Let $S_{\Acal, \Qcal}$ be the cross-similarity matrix between the items in sets $\Acal$ and $\Qcal$. We construct a similarity matrix $S^{\eta,\nu}$ (on a base matrix $S$) in such a way that the cross-similarity between $\Acal$ and $\Qcal$ is multiplied by $\eta$ ({\em i.e.}, $S^{\eta,\nu}_{\Acal,\Qcal} = \eta S_{\Acal,\Qcal}$) to control the trade-off between query-relevance and diversity. Similarly, the cross-similarity between $\Acal$ and $\Pcal$ by $\nu$ ({\em i.e.}, $S^{\eta,\nu}_{\Acal,\Pcal} = \nu S_{\Acal,\Pcal}$) to control the strictness of privacy constraints. Higher values of $\nu$ ensure stricter privacy constraints, such as in the context of privacy-preserving summarization, by tightening the extent of dissimilarity of the subset from the private set. Given the standard form of LogDet as $f(\Acal) = \log\det(S^{\eta,\nu}_{\Acal})$, we provide the MI, CG and CMI expressions in Table~\ref{tab:expr}. For simplicity of notation, CMI is presented with $\nu = \eta = 1$.

\subsection{Facility Location (FL) Instantiations} 

We present two variants of the MI functions for the FL function which is defined as: $f(\Acal) = \sum_{i \in \Omega} \max_{j \in \Acal} S_{ij}$. The first variant is defined over $\Vcal$ (FL\textbf{V}MI) ~\citep{iyer2021submodular}, in Table~\ref{tab:expr}. We derive another variant defined over $\Qcal$ (FL\textbf{Q}MI) which considers only cross-similarities between data points and the target. This MI expression has interesting characteristics different from those of \textsc{Flvmi}. In particular, whereas \textsc{Flvmi} gets saturated ({\em i.e.}, once the query is satisfied, there is no gain in picking another query-relevant data point), \textsc{Flqmi} just models the pairwise similarities of target to data points and vice versa. Moreover, \textsc{Flqmi} only requires a $\Qcal \times \Vcal$ kernel, which makes it very efficient to optimize. In this case, the expression for CG and CMI don't make sense, since they require computing terms over $\Vcal^{\prime}$, which we do not have access to. We multiply the similarity kernel $S$ used in MI and CG expressions of FL by $\eta$ and $\nu$ as done in the case of LogDet.

\subsection{Concave Over Modular (COM)} 

Define a set function $f_{\eta}(\Acal)$ as: 
\begin{align*}
f_{\eta}(\Acal) = &\eta \sum_{i \in \Vcal^{\prime}} \max(\psi(\sum_{j \in \Acal \cap \Vcal} S_{ij}), \psi(\sqrt{n}\sum_{j \in \Acal \cap \Vcal^{\prime}} S_{ij})) \nonumber \\ &+  \sum_{i \in \Vcal} \max(\psi
    (\sum_{j \in \Acal \cap \Vcal^{\prime}} S_{ij}), \psi(\sqrt{n}\sum_{j \in \Acal \cap \Vcal} S_{ij})), 
\end{align*}
where $\psi$ is a concave function and $f_{\eta}(\Acal)$ is restricted submodular. Note that the expression for CG and CMI don't make sense in COM since they require computing terms over $\Vcal^{\prime}$, which we do not have access to.

\subsection{Graph Cut (GC) Instantiations} 

The GC function is defined as $f(\Acal) = \sum \limits_{i \in \Acal, j \in \Vcal} S_{ij} - \lambda \sum\limits_{i, j \in \Acal} S_{ij}$. $S_{ij}$ measure the similarity between elements $i$ and $j$ and the parameter $\lambda$ captures the trade-off between diversity and representativeness. The MI, CG and CMI expressions of GC are presented in Table~\ref{tab:expr}. Note that the CMI expression for GC is not useful as it does not involve the private set and is exactly the same as the MI version. Like in the LogDet case, we introduce an additional parameter $\nu$ in \textsc{Gccg} to control the strictness of privacy constraints. Again, this is easily modeled in the GC objective by multiplying the cross-similarity between data points and the private instances by $\nu$.

\section{Related Work}

To the best of our knowledge, \textsc{Apricot}~\citep{schreiber2020apricot} and \textsc{SFO}~\citep{JMLR:v11:krause10a} are the only other libraries available for submodular optimization. \textsc{SFO} implements algorithms for optimization of submodular functions but is available only as a toolbox to be used in MATLAB or Octave. \textsc{Submodlib} on the other hand is an open-source Python library accessible to a larger community. \textsc{Apricot} is an open-source Python library. However, as compared to \textsc{Apricot}, \textsc{Submodlib} implements a larger suite of functions including the submodular information measures which are not available in \textsc{Apricot}.

\section{\textsc{Submodlib}}

We now present \textsc{Submodlib}, an \href{https://github.com/decile-team/submodlib}{open-source}, easy-to-use, efficient and scalable Python library for submodular optimization with a C++ optimization engine. \textsc{Submodlib} lends itself well to the different applications we have talked about earlier - summarization, data subset selection, hyper parameter tuning, efficient training, guided data subset selection, targeted learning, guided summarization etc. Through a rich API, it offers a great deal of flexibility in the way it can be used.

\subsection{Salient features of \textsc{Submodlib}}

\begin{figure}
    \centering
    \includegraphics[width=\textwidth]{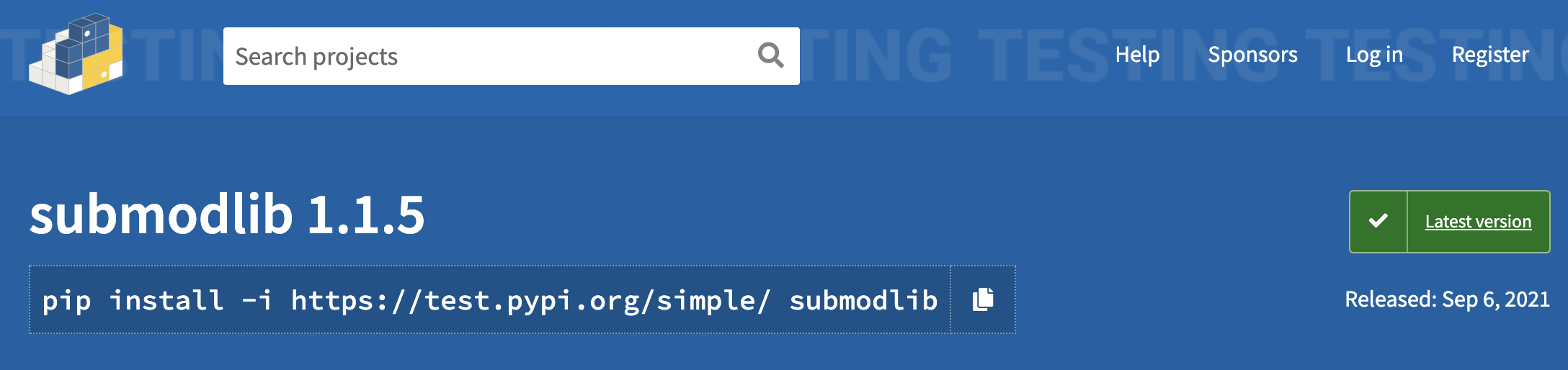}
    \caption{\textsc{Submodlib} is hosted on \textsc{TestPyPi} and is easy to setup}
    \label{fig:testpypi}
\end{figure}

\begin{figure}
    \centering
    \includegraphics[width=\textwidth]{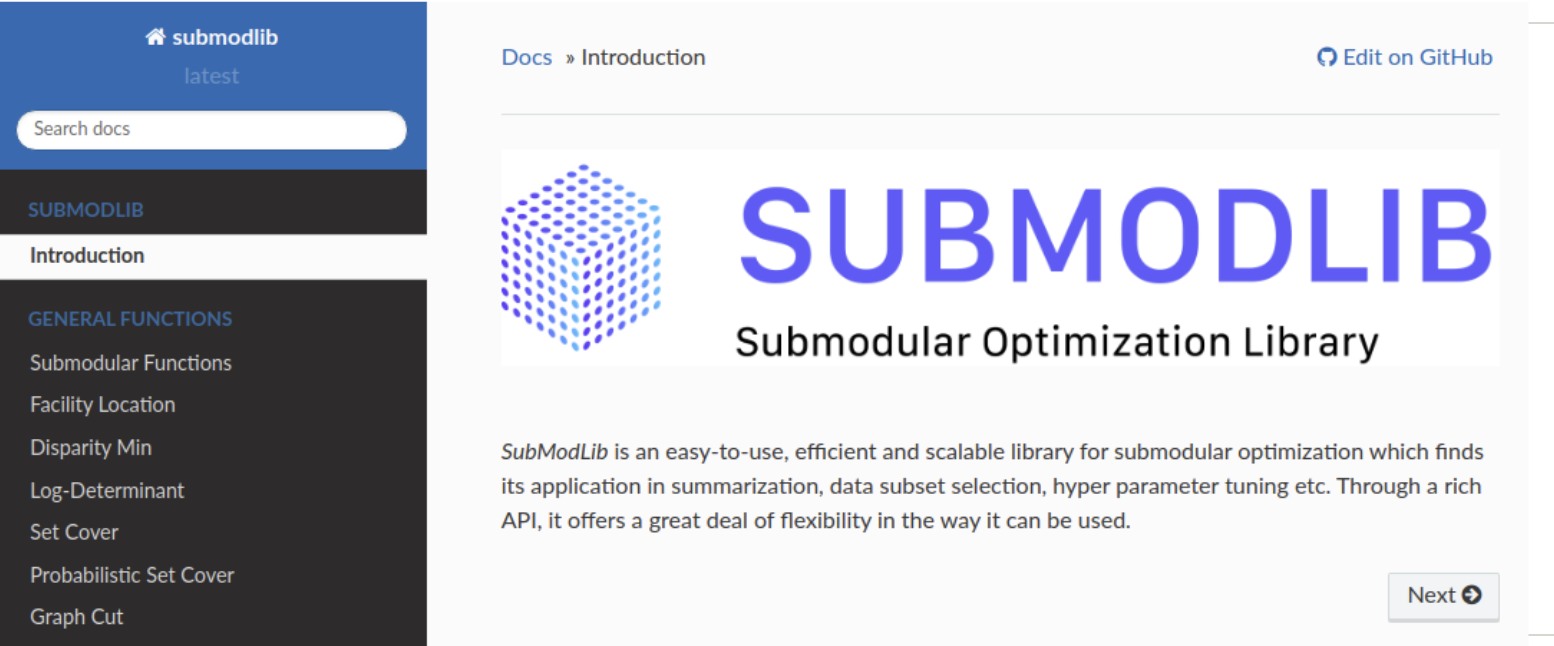}
    \caption{\textsc{Submodlib} comes with a comprehensive documentation hosted on \textsc{ReadTheDocs}}
    \label{fig:doc}
\end{figure}

\textsc{Submodlib} offers an implementation of a rich suite of functions for a wide variety of tasks - regular set (submodular) functions, submodular mutual information functions, conditional gain functions and conditional mutual information functions. \textsc{Submodlib} also supports different types of optimizers - naive greedy, lazy (accelerated) greedy, stochastic (random) greedy and lazier than lazy greedy. It combines the best of Python's ease of use and C++'s efficiency. Through a rich API, \textsc{Submodlib} gives a lot of flexibility and a variety of options to the user, each having a different set of advantages. Further, the de-coupled function and optimizer paradigm (an appropriate function is first instantiated and then maximize() is called on it) makes it suitable for a wide-variety of tasks. \textsc{Submodlib} is hosted on \href{https://test.pypi.org/project/submodlib/}{\textsc{TestPyPi}} and is easy to install with a single `pip install` command (Figure~\ref{fig:testpypi}). It comes with a comprehensive documentation (available at \href{https://submodlib.readthedocs.io/}{\textsc{ReadTheDocs}}) (Figure~\ref{fig:doc}).

\subsection{Functions implemented in \textsc{Submodlib} and implementation notes}

\subsubsection{Regular set (submodular) functions}

As presented in Section~\ref{sec:submod}, these are typically used for vanilla subset selection requiring representation, diversity or coverage.

\textbf{Facility Location: } Facility Location function is implemented as defined earlier. The implementation supports the case when the set whose representation is desired (represented set) is different from the ground set.  

\textbf{Disparity Sum: } Disparity Sum function is implemented as defined earlier.

\textbf{Disparity Min: } Disparity Min function is implemented as defined earlier.

\textbf{Log Determinant: } Log Determinant function is implemented as defined earlier. The implementation leverages Fast Greedy MAP Inference as presented in ~\cite{chen2018fast}.

\textbf{Set Cover: } Set Cover function is implemented as defined earlier.

\textbf{Probabilistic Set Cover: } Probabilistic Set Cover function is implemented as defined earlier.

\textbf{Graph Cut: } Graph Cut function is implemented as defined earlier. The implementation supports the case when the set whose representation is desired (represented set) is different from the ground set. 

\textbf{Feature Based: } Feature-based function is implemented as defined earlier. The implementation supports logarithmic, square root and inverse functions as concave functions. 

\subsubsection{Submodular Mutual Information (MI) Functions}

As introduced in ~\cite{kothawade2021prism}, these are typically used for query-focused subset selection/summarization. Base Mutual Information Function is implemented the way it is defined in Section~\ref{sec:sim}.

\textbf{Facility Location Mutual Information: } Facility Location Mutual Information function is implementation of FLVMI as defined earlier. To avoid duplicate computations during optimization, for each $i \in V$, $\max\limits_{j \in A}s_{ij}$ is maintained as a pre-computed statistic for subset $A$.

\textbf{Facility Location Variant Mutual Information: } Facility Location Variant Mutual Information function is implementation of FLQMI as defined earlier. To avoid duplicate computations during optimization, for each $i \in Q$, $\max\limits_{j \in A}s_{ij}$ is maintained as a pre-computed statistic for subset $A$.

\textbf{Graph Cut Mutual Information: } Graph Cut Mutual Information function is implemented as defined earlier. To avoid duplicate computations during optimization, $\sum \limits_{i \in A} \sum \limits_{j \in Q} s_{ij}$ is maintained as a pre-computed statistic for subset $A$.

\textbf{Log Determinant Mutual Information: } To implement Log Determinant Mutual Information Function, first a Log Determinant function is instantiated with appropriate kernel and then a Mutual Information function is instantiated using it to give Log Determinant Mutual Information Function.

\textbf{Concave Over Modular: } Concave Over Modular function is implemented as defined earlier. To avoid duplicate computations during optimization, for each query element $j \in Q$, $\sum_{i \in \mathcal{A}} s_{ij}$ is maintained as a pre-computed statistic for subset $A$. The implementation supports logarithmic, square root and inverse functions as concave functions. 

\textbf{Set Cover Mutual Information: } A careful examination of the expression of Set Cover Mutual Information function reveals that it is essentially same as Set Cover with cover set of each ground set element (set of concepts covered by that element in the ground set) modified to contain only those concepts which are in the query set. This fact is exploited to implement Set Cover Mutual Information function as a modified Set Cover function.

\textbf{Probabilistic Set Cover Mutual Information: } A careful examination of the expression of Probabilistic Set Cover Mutual Information reveals that it is essentially same as Probabilistic Set Cover with the concept weights set to zero for those concepts which are not present in the query set. This fact is exploited to implement Probabilistic Set Cover Mutual Information function as a modified Probabilistic Set Cover function.

\subsubsection{Conditional Gain (CG) Functions}

These are typically used for query-irrelevant or privacy-preserving subset selection/summarization. Base Conditional Gain Function is implemented the way it is defined in Section~\ref{sec:sim}.

\textbf{Facility Location Conditional Gain: } To implement Facility Location Conditional Gain Function, first a Facility Location function is instantiated with appropriate kernel and then a Conditional Gain function is instantiated using it to give Facility Location Conditional Gain Function.

\textbf{Graph Cut Conditional Gain: } Graph Cut Conditional Gain function is implemented as defined earlier. To avoid duplicate computations during optimization, for each $i \in V$, $\sum_{j \in A} s_{ij}$ is maintained as a pre-computed statistic for subset $A$.

\textbf{Log Determinant Conditional Gain: } To implement Log Determinant Conditional Gain Function, first a Log Determinant function is instantiated with appropriate kernel and then a Conditional Gain function is instantiated using it to give Log Determinant Conditional Gain Function.

\textbf{Set Cover Conditional Gain: } A careful examination of the expression of Set Cover Conditional Gain function reveals that it is essentially same as Set Cover with cover set of each ground set element (set of concepts covered by that element in the ground set) modified to contain only those concepts which are not in the private set. This fact is exploited to implement Set Cover Conditional Gain function as a modified Set Cover function. 

\textbf{Probabilistic Set Cover Conditional Gain: } A careful examination of the expression of Probabilistic Set Cover Conditional Gain reveals that it is essentially same as Probabilistic Set Cover with the concept weights set to zero for those concepts which are present in the private set. This fact is exploited to implement Probabilistic Set Cover Conditional Gain function as a modified Probabilistic Set Cover function.

\subsubsection{Conditional Mutual Information (CMI) Functions: }

These are typically used for joint query-focused and privacy-preserving subset selection/summarization. Base Conditional Mutual Information Function is implemented the way it is defined in Section~\ref{sec:sim}.

\textbf{Facility Location Conditional Mutual Information: } To implement Facility Location Conditional Mutual Information, first a Facility Location function is instantiated with appropriate kernel. A Conditional Gain function is then instantiated using it and finally a Mutual Information function is instantiated using the instance of the Conditional Gain Function.

\textbf{Log Determinant Conditional Mutual Information: } To implement Log Determinant Conditional Mutual Information, first a Log Determinant function is instantiated with appropriate kernel. A Conditional Gain function is then instantiated using it and finally a Mutual Information function is instantiated using the instance of the Conditional Gain Function.

\textbf{Set Cover Conditional Mutual Information: } A careful examination of the expression of Set Cover Conditional Mutual Information reveals that it is essentially same as Set Cover with cover set of each ground set element (set of concepts covered by that element in the ground set) modified to contain only those concepts which are in the query set and not in the private set. This fact is exploited to implement Set Cover Conditional Mutual Information as a modified Set Cover function. 

\textbf{Probabilistic Set Cover Conditional Mutual Information: } A careful examination of the expression of Probabilistic Set Cover Conditional Mutual Information reveals that it is essentially same as Probabilistic Set Cover with the concept weights set to zero for those concepts which are either not present in the query set or are present in the private set. This fact is exploited to implement Probabilistic Set Cover Conditional Mutual Information as a modified Probabilistic Set Cover function. 

We present a summary of expressions of all functions and their parameterizations available in \textsc{Submodlib} in Table~\ref{tab:expr}

\begin{table*}[h!t]
\centering
\begin{tabular}{|p{0.05\textwidth}|p{0.15\textwidth}|p{0.2\textwidth}|p{0.2\textwidth}|p{0.27\textwidth}|} 
\hline \hline
\scriptsize{Name} & \scriptsize{$f^{\theta}(\Acal)$} & \scriptsize{Mutual Information (MI) $I_f^{\theta}(\Acal;\Qcal)$} & \scriptsize{Conditional Gain (CG) $f^{\theta}(\Acal|\Pcal)$} & \scriptsize{Conditional Mutual Information (CMI) $I_f^{\theta}(\Acal;\Qcal|\Pcal)$} \\ \hline 
\hline
\scriptsize{Set Cover (SC)} &  
\scriptsize{$w(\Gamma(\Acal))$} & \scriptsize{SCMI: $w(\Gamma(\Acal) \cap \Gamma(\Qcal))$} & 
\scriptsize{SCCG: $w(\Gamma(\Acal) \setminus \Gamma(\Pcal))$} & 
\scriptsize{SCCMI: $w(\Gamma(\Acal) \cap \Gamma(\Qcal) \setminus \Gamma(\Pcal))$} \\
\hline
\scriptsize{Prob. Set Cover (PSC)} &  
 \scriptsize{$\sum\limits_{i \in \Ucal} w_i \bar{P_i}(\Acal)$} & 
\scriptsize{PSCMI: $\sum\limits_{i \in \Ucal} w_i \bar{P_i(\Acal)} \bar{P_i}(\Qcal)$} & 
\scriptsize{PSCCG: $\sum\limits_{i \in \Ucal} w_i\bar{P_i}(\Acal)P_i(\Pcal)$} &  
\scriptsize{PSCCMI: $\sum\limits_{i \in \Ucal} w_i \bar{P_i}(\Acal)\bar{P_i}(\Qcal)P_i(\Pcal)$} \\ 
\hline
\scriptsize{Graph Cut (GC)} &  
\scriptsize{$\sum \limits_{i \in \Acal, j \in \Vcal} S_{ij} - \lambda \sum\limits_{i, j \in \Acal} S_{ij}$} & \scriptsize{\textsc{Gcmi}: $2\lambda \sum\limits_{i \in \Acal} \sum\limits_{j \in \Qcal} S_{ij}$} & 
\scriptsize{\textsc{Gccg}: $f_{\lambda}(\Acal) - 2 \lambda \nu \sum\limits_{i \in \Acal, j \in \Pcal} S_{ij}$} &  
\scriptsize{GCCMI: $2\lambda \sum\limits_{i \in \Acal} \sum\limits_{j \in \Qcal} S_{ij}$} \\ 
\hline
\scriptsize{Log Determinant (LogDet)} &  
\scriptsize{$\log\det(S_{\Acal})$} & 
\scriptsize{\textsc{Logdetmi}: $\log\det(S_{\Acal}) -\log\det(S_{\Acal} - \eta^2 S_{\Acal,\Qcal}S_{\Qcal}^{-1}S_{\Acal,\Qcal}^T)$} & 
\scriptsize{\textsc{Logdetcg}: $\log\det(S_{\Acal} - \nu^2  S_{\Acal,\Pcal}S_{\Pcal}^{-1}S_{\Acal,\Pcal}^T)$} & 
\scriptsize{\textsc{Logdetcmi}: $\log \frac{\det(I - S_{\Pcal}^{-1}S_{\Pcal, \Qcal} S_{\Qcal}^{-1}S_{\Pcal, \Qcal}^T)}{\det(I - S_{\Acal \Pcal}^{-1} S_{\Acal \Pcal, Q} S_{\Qcal}^{-1} S_{\Acal \Pcal, Q}^T)}$} \\ 
\hline
\scriptsize{Facility Location (FL) (v1)} &  
\scriptsize{$\sum\limits_{i \in \Vcal} \max\limits_{j \in \Acal} S_{ij}$} &
\scriptsize{FLVMI: $\sum\limits_{i \in \Vcal}\min(\max\limits_{j \in \Acal}S_{ij}, \eta \max\limits_{j \in \Qcal}S_{ij})$} & 
\scriptsize{\textsc{Flcg}: $\sum\limits_{i \in \Vcal} \max(\max\limits_{j \in \Acal} S_{ij}-$ $\nu \max\limits_{j \in \Pcal} S_{ij}, 0)$} & 
\scriptsize{\textsc{Flcmi}: $\sum\limits_{i \in \Vcal} \max(\min(\max\limits_{j \in \Acal} S_{ij},$ $\eta \max\limits_{j \in \Qcal} S_{ij}) - \nu \max\limits_{j \in \Pcal} S_{ij}, 0)$} \\
\hline
\scriptsize{Facility Location (FL) (v2)} &  
\scriptsize{$\sum\limits_{i \in \Omega} \max\limits_{j \in \Acal} S_{ij}$} & 
\scriptsize{\textsc{Flqmi}: $\sum\limits_{i \in \Qcal} \max\limits_{j \in \Acal} S_{ij} + $ $\eta \sum\limits_{i \in \Acal} \max\limits_{j \in \Qcal} S_{ij}$} & 
\scriptsize{FL2CG: Not Useful} & 
\scriptsize{FL2CMI: Not Useful} \\
\hline
\scriptsize{Concave Over Modular (COM)} & 
\scriptsize{See text}
 & 
\scriptsize{$\eta \sum_{i \in \Acal} \psi(\sum_{j \in \Qcal}S_{ij}) + \sum_{j \in \Qcal} \psi(\sum_{i \in \Acal} S_{ij})$} & 
\scriptsize{Not Useful} & \scriptsize{Not Useful} \\
\hline
\hline
\end{tabular}
\caption{Suite of functions implemented in \textsc{Submodlib}. As discussed in ~\cite{kothawade2021prism} some are not particularly useful}
\label{tab:expr}
\end{table*}

\subsection{Optimizers implemented in \textsc{Submodlib}}

\subsubsection{Naive Greedy}

Given a set of items $V = \{1, 2, 3, \cdots, n\}$ which we also call the \emph{Ground Set}, define a utility function (set function) $f:2^V \rightarrow \Re$, which measures how good a subset $A \subseteq V$ is. Let $c :2^V \rightarrow \Re$ be a cost function, which describes the cost of the set (for example, the size of the subset). The goal is then to have a subset $A$ which maximizes $f$ while simultaneously minimizing the cost function $c$. It is easy to see that maximizing a generic set function becomes computationally infeasible as $V$ grows. Often the cost $c$ is budget constrained (for example, a fixed set summary) and a natural formulation of this is the following problem:

$$\max\{f(A) \mbox{ such that } c(A) \leq b\}$$

The naive greedy optimizer implementation in \textsc{Submodlib} implements the standard greedy algorithm ~\citep{minoux1978accelerated}. It starts with an empty set and in every iteration adds to it a new element from the ground set with maximum marginal gain until the desired budget is achieved or the best gain in any iteration is zero or negative. The solution thus obtained is called a \emph{greedy solution}. 

When $f$ is a submodular function, using a simple greedy algorithm to compute the above gives a lower-bound performance guarantee of around 63\% of optimal ~\citep{nemhauser1978analysis} and in practice these greedy solutions are often within 90\% of optimal ~\citep{krause2008optimizing}.

It is important to note that unless the marginal gain of the element at each step is unique, the greedy solution will not be unique. In such a case, the current implementation adds the first best element encountered at every iteration. As unordered sets are used to represent the ground sets, this ordering need not be unique.

\subsubsection{Lazy Greedy or Accelerated Greedy}

The lazy greedy optimizer in \textsc{Submodlib} is an implementation of the accelerated greedy algorithm described in ~\cite{minoux1978accelerated}. Essentially, it maintains an upper bound of the marginal gain of every item and reduces them as the optimal set grows. Due to the submodularity of the function, it is guaranteed that the marginal gain of any element on a set will always be less than or equal to that on a smaller set. In any iteration, because of maintaining the upper bounds in a descending order, the algorithm doesn't have to scan the entire remaining ground set to look for the next best element to add. Thus lazy greedy optimizer is several times faster than the naive greedy optimizer. The best element is added in every iteration until the desired budget is achieved or the best gain in any iteration is zero or negative.

Since the algorithm exploits submodularity of the function, LazyGreedy optimizer will work only for functions that are guaranteed to be submodular.

\subsubsection{Stochastic (Random) Greedy}

The stochastic greedy optimizer is an implementation of the stochastic greedy algorithm proposed by ~\cite{mirzasoleiman2015lazier}. The main idea is to improve over naive greedy by a sub-sampling step. Specifically, in each step it first samples a set $R$ of size $(n/k)\log(1/\epsilon)$ uniformly at random and then adds that element from $R$ to the greedy set $A$ which increases its value the most. Such an element is added in every iteration until the desired budget is achieved or the best gain in any iteration is zero or negative.

Stochastic greedy optimizer has provably linear running time independent of the budget, while simultaneously having the same approximation ratio guarantee (in expectation). It is substantially faster than both naive greedy and lazy greedy optimizers.

Also, at a very high level stochastic greedy's improvement over naive greedy is similar in spirit to how stochastic gradient descent improves the running time of gradient descent for convex optimization. 

\subsubsection{Lazier Than Lazy Greedy}

The implementation of lazier-than-lazy greedy optimizer in \textsc{Submodlib} is an implementation of "random sampling with lazy evaluation" proposed by ~\cite{mirzasoleiman2015lazier}. It combines both stochastic greedy and lazy greedy approaches. Essentially, in every iteration, it applies lazy greedy for finding the best element from a random sub sample of the remaining ground set and adds that element to the greedy set. Such an element is added in every iteration until the desired budget is achieved or the best gain in any iteration is zero or negative.

For submodular functions, LazierThanLazyGreedy optimizer is the most efficient, followed by StochasticGreedy, LazyGreedy and NaiveGreedy in the descending order of speed. We demonstrate this empirically in the following section. 

\subsubsection{Comparison of different optimizers}

\begin{figure}
    \centering
    \includegraphics{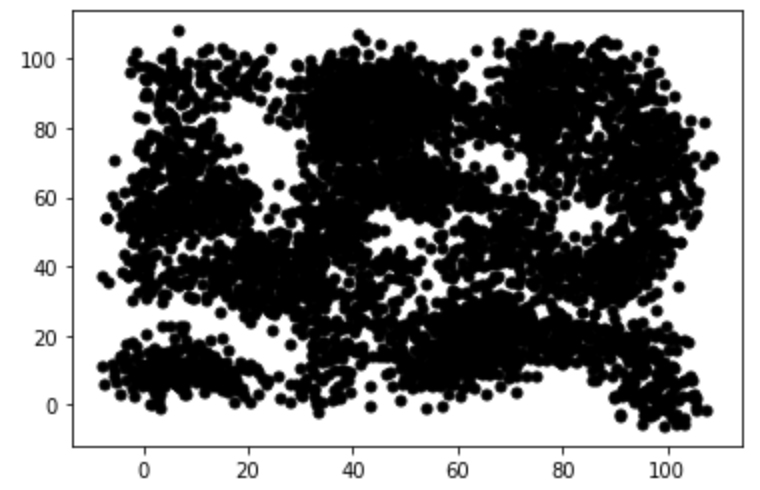}
    \caption{Dataset for comparing the performance of different optimizers}
    \label{fig:optimizer-dataset}
\end{figure}

We create a synthetic dataset of 500 points distributed across 10 clusters with a standard deviation of 4. In practice, each data point could correspond to images or video frames or any other dataset where subset selection is desired. Figure~\ref{fig:optimizer-dataset} visualizes this dataset. 

We run the following code to measure the running time of different optimizers and present the numbers in Table~\ref{tab:numbers}.

\begin{verbatim}
    %timeit greedyList = obj.maximize(budget=10, optimizer='LazyGreedy', stopIfZeroGain=False, 
                         stopIfNegativeGain=False, verbose=False)
\end{verbatim}

\begin{table}[]
\centering
\begin{tabular}{|l|l|}
\hline
\textbf{Optimizer} & \textbf{Running Time} \\ \hline
Naive Greedy            & 1 loop, best of 5: 3.93 s per loop \\ \hline
Stochastic Greedy       & 1 loop, best of 5: 1.17 s per loop \\ \hline
Lazy Greedy             & 1 loop, best of 5: 417 ms per loop \\ \hline
Lazier Than Lazy Greedy & 1 loop, best of 5: 405 ms per loop \\ \hline
\end{tabular}
\caption{Comparison of running times of different optimizers}
\label{tab:numbers}
\end{table}





\section{Efficient optimization in \textsc{Submodlib}}

The computational complexity of optimizing the above functions can be further reduced by pre-computing certain statistics (called memoization) so as to avoid repeated/redundant computations through the iterations of the above greedy approaches. The memoization implemented in \textsc{Submodlib} for the regular functions are as presented in Table~\ref{tab:memoization-existing} and the memoization for other functions are as discussed above. We summarize them in Table~\ref{tab:other-memoization}. This makes \textsc{Submodlib} especially efficient.

\begin{table*}
\begin{center}
 \begin{tabular}{| c | c | c |} 
 \hline
 \textbf{Function} & \textbf{Expression} & \textbf{Pre-compute Statistic} \\ [0.5ex] 
 \hline
 Facility Location & $\sum_{i \in V} \max_{k \in A} s_{ik}$ & $[\max_{k \in A} s_{ik}, i \in V]$ \\ 
 \hline
 Graph Cut & $\lambda \sum_{i \in V}\sum_{j \in A} s_{ij} - \sum_{i, j \in A} s_{ij}$ & $[\sum_{j \in A} s_{ij}, i \in V]$ \\ 
\hline
 Feature Based & $\sum_{i \in \mathcal F} \psi(w_i(A))$ & $[w_i(A), i \in \mathcal F]$  \\
 \hline
 Set Cover & $w(\cup_{i \in A} U_i)$ & $\cup_{i \in A} U_i$ \\
 \hline
 Prob. Set Cover & $\sum_{i \in \mathcal U} w_i[1 - \prod_{k \in A}(1 - p_{ik})]$ & $[\prod_{k \in A} (1 - p_{ik}), i \in \mathcal U]$  \\ [1ex] 
 \hline
 DPP & $\log\det(S_A))$ & SVD($S_A$) \\
 \hline
 Dispersion Min & $\min_{k,l  \in A, k \neq l} d_{kl}$ & $\min_{k, l \in A, k \neq l} d_{kl}$ \\
 \hline
 Dispersion Sum & $\sum_{k,l  \in A} d_{kl}$ & $[\sum_{k \in A} d_{kl}, l \in A]$ \\
  \hline
 \hline
\end{tabular}
\caption{Memoization implemented in \textsc{Submodlib} for regular functions.}
\label{tab:memoization-existing}
\end{center}
\end{table*}

\begin{table}[]
\small
\centering
\begin{tabular}{|l|l|l|}
\hline
\textbf{Function} & \textbf{Expression}                                                                                                                                                                                                                                                                                                                          & \textbf{Pre-compute Statistic}                                 \\ \hline
FLVMI    & $\sum\limits_{i \in \Vcal}\min(\max\limits_{j \in \Acal}S_{ij}, \eta \max\limits_{j \in \Qcal}S_{ij})$                                                                                                                                                                                                                              & $\max\limits_{j \in A}s_{ij}$, $i \in V$                   \\ \hline
FLQMI    & $\sum\limits_{i \in \Qcal} \max\limits_{j \in \Acal} S_{ij} + $ $\eta \sum\limits_{i \in \Acal} \max\limits_{j \in \Qcal} S_{ij}$                                                                                                                                                                                                   & $\max\limits_{j \in A}s_{ij}$, $i \in Q$                   \\ \hline
GCMI     & $2\lambda \sum\limits_{i \in \Acal} \sum\limits_{j \in \Qcal} S_{ij}$                                                                                                                                                                                                                                                               & $\sum \limits_{i \in A} \sum \limits_{j \in Q} s_{ij}$              \\ \hline
COM      & Refer text & $\sum_{i \in \mathcal{A}} s_{ij}$, $j \in Q$ \\ \hline
GCCG     & $f_{\lambda}(\Acal) - 2 \lambda \nu \sum\limits_{i \in \Acal, j \in \Pcal} S_{ij}$                                                                                  & $\sum_{j \in A} s_{ij}$, $i \in V$                        \\ \hline
\end{tabular}
\caption{Memoization implemented in \textsc{Submodlib} for computational efficiency. As other functions are built on top of existing ones, they reuse their memoization.}
\label{tab:other-memoization}
\end{table}

\section{Sample usage}

It is very easy to get started with \textsc{Submodlib}. Using a submodular function in \textsc{Submodlib} essentially boils down to just two steps:

\begin{enumerate}
    \item instantiate the corresponding function object
    \item invoke the desired method on the created object
\end{enumerate}

The most frequently used methods are:

\begin{enumerate}
    \item f.evaluate() - takes a subset and returns the score of the subset as computed by the function f
    \item f.marginalGain() - takes a subset and an element and returns the marginal gain of adding the element to the subset, as computed by f
    \item f.maximize() - takes a budget and an optimizer to return an optimal set as a result of maximizing f
\end{enumerate}

Thus, a basic usage of \textsc{Submodlib} calls for the following code:

\begin{verbatim}
    from submodlib import FacilityLocationFunction
    objFL = FacilityLocationFunction(n=43, data=groundData, mode="dense", metric="euclidean")
    greedyList = objFL.maximize(budget=10,optimizer='NaiveGreedy')
\end{verbatim}

Next we present other advanced usage options provided by \textsc{Submodlib} that offers a great deal of flexibility in the hands of a user.

\section{Different usage patterns supported by \textsc{Submodlib}}

\textsc{Submodlib} provides different alternative ways of consumption based on user requirements and or scale/efficiency reasons. For example, subset selection by Facility Location can be invoked in following different ways:

\begin{enumerate}
    \item Creating dense similarity kernel in C++: user only provides the data matrix and dense similarity kernel is internally created in C++ using user specified similarity metric
    \item Creating dense similarity kernel in Python: user creates dense similarity kernel in Python (using \textsc{Submodlib}'s helper code) and uses that to instantiate FacilityLocation
\end{enumerate}

For both of the above versions, user can opt for creating sparse similarity kernel (similarity with points beyond the num\_neighbors is considered zero) as against dense similarity kernel (N X N). Sparse kernels tend to be more efficient, especially for large datasets, but at the cost of accuracy.

In addition to the above, as another alternative for efficient and scalable implementation and to provide for supervised subset selection, \textsc{Submodlib} provides clustered implementation of various submodular functions. \textsc{Submodlib} does this in two ways:

\begin{enumerate}
    \item As yet another "mode" in the particular function - for example "clustered" mode in FacilityLocation over and above "dense" and "sparse" modes. An alternative clustered implementation of Facility Location assumes a clustering of all ground set items and then the function value is computed over the clusters as 
	$$f(A) = \sum_{l \in {1....k}} \sum_{i \in C_l} \max_{j \in A \cap C_l} s_{ij}$$ 
    \item Through a generic \textbf{Clustered Function} implementation which works for any submodular function. Given a set-function $f$ and a clustering, clustered function internally creates a mixture of functions each defined over a cluster. It is thus defined as
	
	$$f(A) = \sum_i f_{C_i}(A)$$
	
	where $f_{C_i}$ operates only on cluster $C_i$ as sub-groundset and interprets $A$ as $A \cap C_i$
\end{enumerate} 

For both the alternatives, and in the spirit of giving flexibility, the user has choice to either a) let \textsc{Submodlib} do the clustering internally or b) provide the clusters (for example in case of supervised subset selection).

\section{Timing analysis of \textsc{Submodlib}}

To gauge the performance of \textsc{Submodlib}, selection by Facility Location was performed on a randomly generated dataset of 1024-dimensional points. Specifically the following code was run for the number of data points ranging from 50 to 10000.

\begin{verbatim}
    K_dense = helper.create_kernel(dataArray, mode="dense", 
              metric='euclidean', method="other")
    obj = FacilityLocationFunction(n=num_samples, mode="dense", sijs=K_dense, 
          separate_rep=False, pybind_mode="array")
    obj.maximize(budget=budget,optimizer=optimizer, stopIfZeroGain=False, 
                stopIfNegativeGain=False, verbose=False, show_progress=False)
\end{verbatim}

The above code was timed using Python's \textsc{timeit} module averaged across three executions each. We report the numbers in Table~\ref{tab:submodlib-timing}.

\begin{table}[]
\centering
\begin{tabular}{|
>{\columncolor[HTML]{FFFFFF}}l |
>{\columncolor[HTML]{FFFFFF}}l |}
\hline
\multicolumn{1}{|c|}{\cellcolor[HTML]{FFFFFF}{\color[HTML]{24292F} \textbf{Number of data points}}} & \multicolumn{1}{c|}{\cellcolor[HTML]{FFFFFF}{\color[HTML]{24292F} \textbf{Time taken (in seconds)}}} \\ \hline
{\color[HTML]{24292F} 50}                                                                           & {\color[HTML]{24292F} 0.00043}                                                                       \\ \hline
{\color[HTML]{24292F} 100}                                                                          & {\color[HTML]{24292F} 0.001074}                                                                      \\ \hline
{\color[HTML]{24292F} 200}                                                                          & {\color[HTML]{24292F} 0.003024}                                                                      \\ \hline
{\color[HTML]{24292F} 500}                                                                          & {\color[HTML]{24292F} 0.016555}                                                                      \\ \hline
{\color[HTML]{24292F} 1000}                                                                         & {\color[HTML]{24292F} 0.081773}                                                                      \\ \hline
{\color[HTML]{24292F} 5000}                                                                         & {\color[HTML]{24292F} 2.469303}                                                                      \\ \hline
{\color[HTML]{24292F} 6000}                                                                         & {\color[HTML]{24292F} 3.563144}                                                                      \\ \hline
{\color[HTML]{24292F} 7000}                                                                         & {\color[HTML]{24292F} 4.667065}                                                                      \\ \hline
{\color[HTML]{24292F} 8000}                                                                         & {\color[HTML]{24292F} 6.174047}                                                                      \\ \hline
{\color[HTML]{24292F} 9000}                                                                         & {\color[HTML]{24292F} 8.010674}                                                                      \\ \hline
{\color[HTML]{24292F} 10000}                                                                        & {\color[HTML]{24292F} 9.417298}                                                                      \\ \hline
\end{tabular}
\caption{Timing analysis of \textsc{Submodlib}}
\label{tab:submodlib-timing}
\end{table}

\section{Sample applications using \textsc{Submodlib}}

Below we demonstrate the usage of \textsc{Submodlib} in some example scenarios.

\subsection{Using \textsc{Submodlib} to study the modeling capabilities of different submodular functions}

\begin{figure}[H]
\centering
    \includegraphics[width=0.6\textwidth]{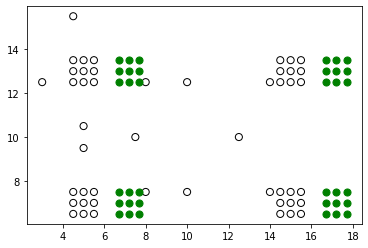}
    \caption{Synthetic dataset to study the behavior of different functions in selecting a subset under maximization. Hollow points constitute the ground set and green points constitute the set whose representation may be desired (represented set)}
    \label{fig:data}
\end{figure}

To carefully observe the characteristics of the optimal set obtained by the maximization of different functions, we begin by creating a controlled dataset of 48 2D points as shown by the hollow circles in Figure~\ref{fig:data}. Specifically we have some clusters and some outliers in this dataset. We also create a different set (green points) whose representation may be desired. This is to demonstrate that this set need not be same as the ground set (whose subset is desired).

\begin{figure*}[h]
\centering
\begin{subfigure}{0.4\textwidth}
    \includegraphics[width=\linewidth]{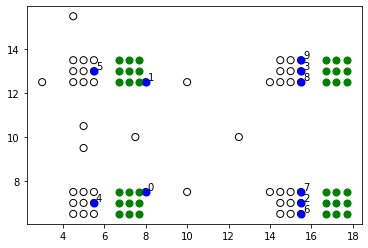}
    \caption{Selection by FacilityLocation}
    \end{subfigure}
    \begin{subfigure}{0.4\textwidth}
    \includegraphics[width=\linewidth]{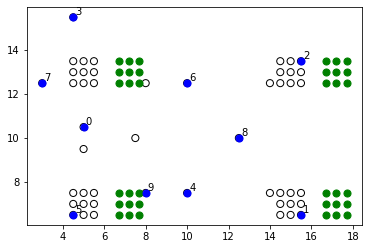}
    \caption{Selection by DisparitySum}
    \end{subfigure}
    \caption{Selection of subset (blue circles) by a) Facility Location and b) Disparity-Sum function. The subset selected by FacilityLocation is clearly representative of the represented set, while the subset selected by DisparitySum contains the outliers in order to favor diversity. The labels indicate the order in which the points got selected during greedy maximization}
    \label{fig:fl-dsum}
\end{figure*}

As an illustrative example, we compare the behavior of FacilityLocation and DisparitySum functions in selecting optimal subset. We first maximize Facility Location function to find the optimal set of size 10. We visualize the points in the optimal set (blue)(Figure~\ref{fig:fl-dsum}(a)). The selected points are numbered in the order in which they get picked up by the naive greedy max algorithm.

\begin{verbatim}
    from submodlib import FacilityLocationFunction
    objFL = FacilityLocationFunction(n=48, data=groundData, separate_rep=True, 
            n_rep=36, data_rep=repData, mode="dense", metric="euclidean")
    greedyList = objFL.maximize(budget=10,optimizer='NaiveGreedy', stopIfZeroGain=False, 
                 stopIfNegativeGain=False, verbose=False)
    greedyXs = [groundxs[x[0]] for x in greedyList]
    greedyYs = [groundys[x[0]] for x in greedyList]
    plt.scatter(groundxs, groundys, s=50, facecolors='none', edgecolors='black', label="Images")
    plt.scatter(repxs, repys, s=50, color='green', label="Images")
    plt.scatter(greedyXs, greedyYs, s=50, color='blue', label="Greedy Set")
    for label, element in enumerate(greedyList):
        plt.annotate(label, (groundxs[element[0]],
                    groundys[element[0]]), (groundxs[element[0]]+0.1, groundys[element[0]]+0.1))
\end{verbatim}

We observe that the cluster centers get picked up first followed by the other diverse points in the data set and the outlier point is picked up only at the end. Had the budget been less than 10, the outlier wouldn't even have got picked up. This is expected as Facility Location primarily models representation.

Next we maximize Disparity Sum function to find the optimal set of size 10. We visualize the points in the optimal set (blue)(Figure~\ref{fig:fl-dsum}(b)). The selected points are numbered in the order in which they get picked up by the naive greedy max algorithm.

\begin{verbatim}
    from submodlib import DisparitySumFunction
    objDM = DisparitySumFunction(n=48, data=groundData, mode="dense", metric="euclidean")
    greedyList = objDM.maximize(budget=10,optimizer='NaiveGreedy', stopIfZeroGain=False, 
                 stopIfNegativeGain=False, verbose=False)
    greedyXs = [groundxs[x[0]] for x in greedyList]
    greedyYs = [groundys[x[0]] for x in greedyList]
    plt.scatter(groundxs, groundys, s=50, facecolors='none', edgecolors='black', label="Images")
    plt.scatter(repxs, repys, s=50, color='green', label="Images")
    plt.scatter(greedyXs, greedyYs, s=50, color='blue', label="Greedy Set")
    for label, element in enumerate(greedyList):
        plt.annotate(label, (groundxs[element[0]], groundys[element[0]]), (groundxs[element[0]]+0.1,
                    groundys[element[0]]+0.1))
\end{verbatim}

In case of Disparity Sum we observe that the remote corner points get picked up first followed by the other diverse points in the data set including the outlier point. This is expected as Disparity Sum primarily models diversity and wouldn't mind picking up outlier points if that makes the set diverse enough, compromising on the representativeness of the set with respect to the ground set.

\subsubsection{Using \textsc{Submodlib} to study the modeling capabilities of different submodular mutual information functions}

\begin{figure}[H]
\centering
    \includegraphics[width=0.7\textwidth]{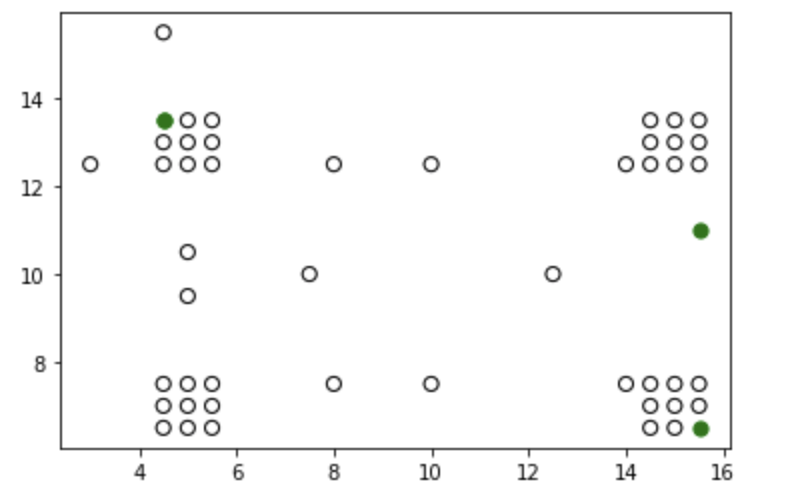}
    \caption{Synthetic dataset to study the behavior of different submodular mutual information (MI) functions in selecting a subset under maximization. Hollow points constitute the ground set and green points constitute the universe set of queries}
    \label{fig:midata}
\end{figure}

To carefully observe the characteristics of the optimal set obtained by the maximization of different submodular mutual information (MI) functions, we begin by creating a controlled dataset of 46 2D points (hollow circles in Figure~\ref{fig:midata}) along with some query points (green circles in Figure~\ref{fig:midata}). Specifically we have some clusters and some outliers in this dataset. Please note that there is no overlap between the ground set and the query set.

\begin{figure*}[h]
    \includegraphics[width=\textwidth]{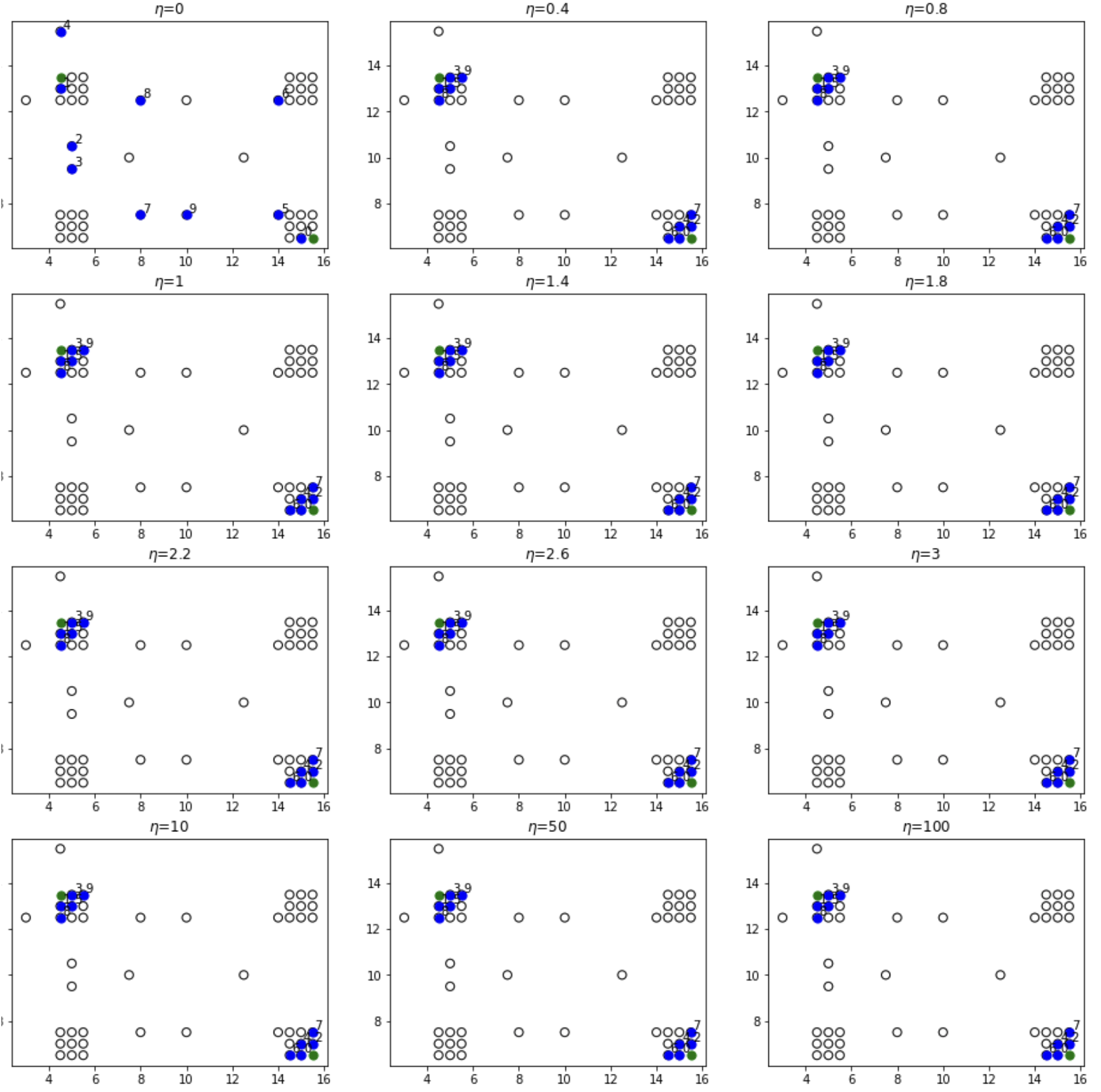}
    \caption{Selection of subset by FLQMI function (blue points) for the given query/target set (green circles) for different values of $\eta$}
    \label{fig:flqmi}
\end{figure*}

We first maximize FLQMI to compute the optimal subset targeted to the given query set. To observe the effect of $\eta$ we do the selections for different values of $\eta$. We show the selections in Figure~\ref{fig:flqmi}.

\begin{verbatim}
    from submodlib import FacilityLocationVariantMutualInformationFunction
    etas = [0, 0.4, 0.8, 1, 1.4, 1.8, 2.2, 2.6, 3, 10, 50, 100]
    row = 0
    index = 1
    plt.figure(figsize = (16, 16))
    for eta in etas:
        plt.subplot(4,3,row*3+index)
        obj = FacilityLocationVariantMutualInformationFunction(n=46, num_queries=2, data=groundData, 
              queryData=mutlipleQueryData, metric="euclidean", queryDiversityEta=eta)
        greedyList = obj.maximize(budget=10,optimizer='NaiveGreedy', stopIfZeroGain=False, 
                    stopIfNegativeGain=False, verbose=False)
        greedyXs = [groundxs[x[0]] for x in greedyList]
        greedyYs = [groundys[x[0]] for x in greedyList]
        plt.scatter(groundxs, groundys, s=50, facecolors='none', edgecolors='black', label="Images")
        plt.scatter(multiplequeryxs, multiplequeryys, s=50, color='green', label="Queries")
        plt.scatter(greedyXs, greedyYs, s=50, color='blue', label="Greedy Set")
        for label, element in enumerate(greedyList):
            plt.annotate(label, (groundxs[element[0]], groundys[element[0]]), (groundxs[element[0]]+0.1,
                        groundys[element[0]]+0.1))
        plt.title('$\eta$='+str(eta))
        index += 1
        if index == 4:
            row += 1
            index = 1
\end{verbatim}

We see that at $\eta$=0, \textsc{Flqmi} picks one query-relevant point each and saturates. It becomes highly query-relevant thereafter. \textsc{Flqmi} tends to have low query-coverage. Higher $\eta$ reduces query-coverage even further.

\begin{figure*}[h]
\centering
    \includegraphics[width=0.6\textwidth]{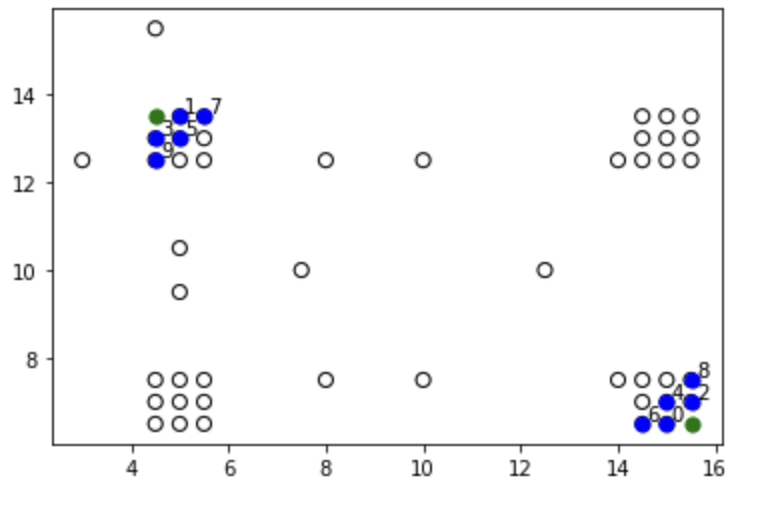}
    \caption{Selection of subset by GCMI function (blue points) for the given query/target set (green circles)}
    \label{fig:gcmi}
\end{figure*}

To contrast the behavior of \textsc{Flqmi} with \textsc{Gcmi}, next we maximize \textsc{Gcmi} to get the same selection. Note that as per the formulation of \textsc{Gcmi}, it doesn't depend on $\eta$. We show the selection in Figure~\ref{fig:gcmi}.

\begin{verbatim}
    from submodlib import GraphCutMutualInformationFunction
    obj = GraphCutMutualInformationFunction(n=46, num_queries=2, data=groundData,
          queryData=mutlipleQueryData, metric="euclidean")
    greedyList = obj.maximize(budget=10,optimizer='NaiveGreedy', stopIfZeroGain=False, 
                 stopIfNegativeGain=False, verbose=False)
    greedyXs = [groundxs[x[0]] for x in greedyList]
    greedyYs = [groundys[x[0]] for x in greedyList]
    plt.scatter(groundxs, groundys, s=50, facecolors='none', edgecolors='black', label="Images")
    plt.scatter(multiplequeryxs, multiplequeryys, s=50, color='green', label="Queries")
    plt.scatter(greedyXs, greedyYs, s=50, color='blue', label="Greedy Set")
    for label, element in enumerate(greedyList):
        plt.annotate(label, (groundxs[element[0]], groundys[element[0]]), (groundxs[element[0]]+0.1,
                    groundys[element[0]]+0.1))
\end{verbatim}

We see that \textsc{Gcmi} acts as a pure retrieval function as it is highly query-relevant doesn't consider diversity among the points in the selected subset.

\subsubsection{Using \textsc{Submodlib} on a real-world image collection}

\begin{figure*}[h!]
\centering
\begin{subfigure}{0.4\textwidth}
\centering
\includegraphics[width = \textwidth]{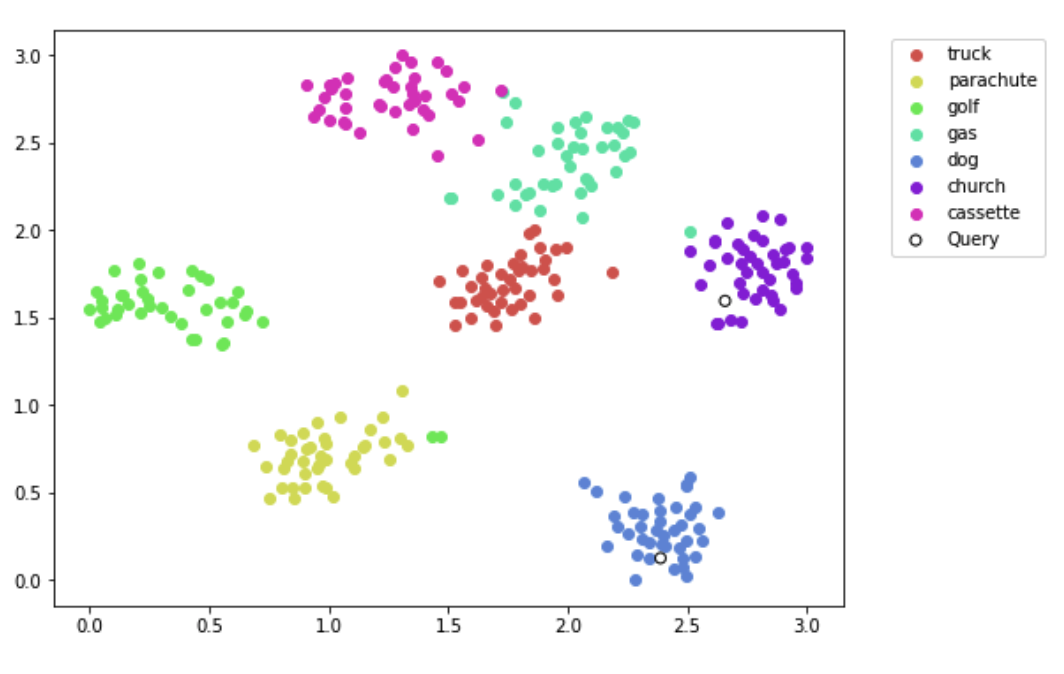}
\caption{Distribution of dataset used}
\end{subfigure}
\begin{subfigure}{0.4\textwidth}
\centering
\includegraphics[width = \textwidth]{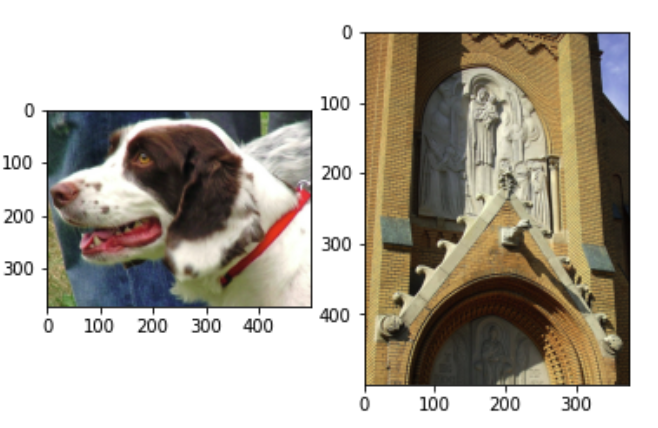}
\caption{The two query images used in the query set}
\end{subfigure}
\caption{a) t-SNE visualization of the dataset, b) the two query images used in the query set.}
\label{fig:imagenette}
\end{figure*}

Finally, we demonstrate the application of \textsc{Submodlib} on a real-world image collection. We use a subset of Imagenette dataset (https://github.com/fastai/imagenette). The dataset along with the query images are visualized using t-SNE in Figure~\ref{fig:imagenette}.

We extract 4096 dimensional VGG fatures for each image in the ground set and for the query images and use them for instantiating the kernels required by the \textsc{Flqmi} function. As before, we then maximize FLQMI to compute the optimal targeted subset aligned with the given query set.

\begin{verbatim}
    from submodlib import FacilityLocationVariantMutualInformationFunction
    obj = FacilityLocationVariantMutualInformationFunction(n=280, num_queries=2, 
          query_sijs=image_query_kernel, metric="cosine")
    greedyList = obj.maximize(budget=20,optimizer='NaiveGreedy', stopIfZeroGain=False, 
                 stopIfNegativeGain=False, verbose=False)
    row = 0
    index = 1
    plt.figure(figsize=(20, 16))
    for elem in greedyList:
      plt.subplot(4, 5, row*5+index)
      image_path = os.path.join(path, classes_id[labels[elem[0]]], ground_images[elem[0]])
      img = cv2.imread(image_path)
      img = cv2.cvtColor(img, cv2.COLOR_BGR2RGB) # cv2 load images as BGR, convert it to RGB
      plt.imshow(img)
      index += 1
      if index == 6:
        row += 1
        index = 1
\end{verbatim}

\begin{figure*}[h!]
\centering
\begin{subfigure}{0.4\textwidth}
\centering
\includegraphics[width = \textwidth]{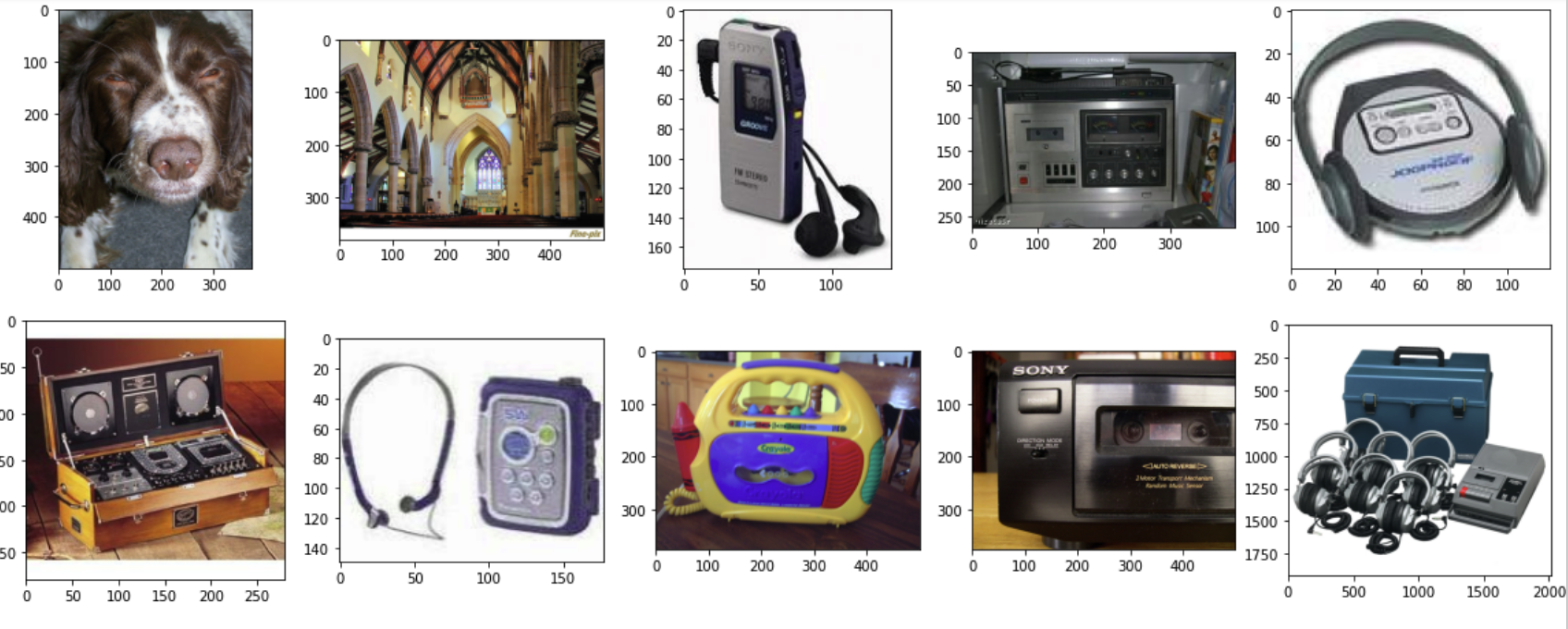}
\caption{}
\end{subfigure}
\begin{subfigure}{0.4\textwidth}
\centering
\includegraphics[width = \textwidth]{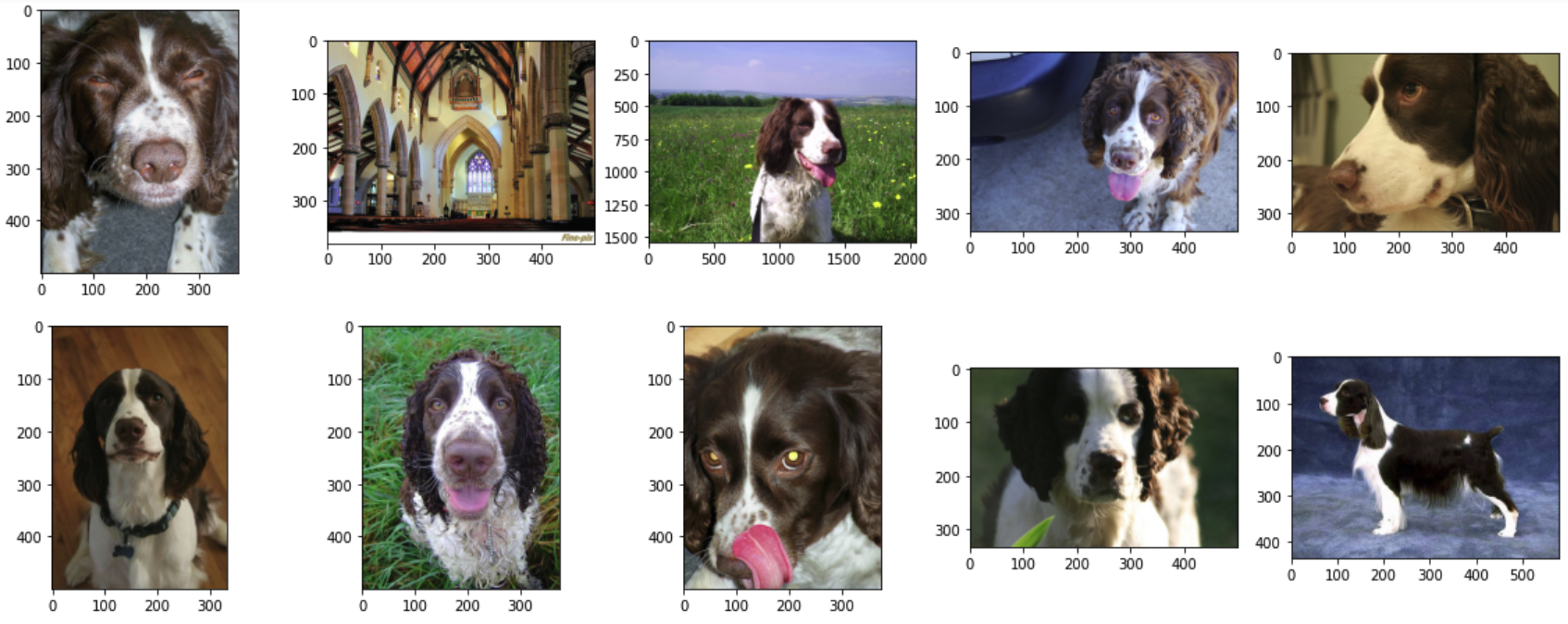}
\caption{}
\end{subfigure}
\caption{a) At $\eta$=0 FLQMI picks one query relevant element for each query and gets saturated, b) Even with a slight increase of $\eta$ (to 0.1) FLQMI becomes highly query relevant, but unfair.}
\label{fig:flqmi-image}
\end{figure*}

The qualitative results for FLQMI are presented in Figure~\ref{fig:flqmi-image}. As seen earlier on the synthetic dataset, at $\eta$=0, FLQMI picks one query-relevant point each and saturates. It becomes highly query-relevant thereafter. FLQMI tends to have low query-coverage. Higher $\eta$ reduces query-coverage even further.

The project README at https://github.com/decile-team/submodlib links to several Google Colab notebooks that demonstrate more usage of \textsc{Submodlib} for various other tasks.

\section{Conclusion}

We presented \textsc{Submodlib}, an open-source, easy-to-use, efficient and scalable Python library for submodular optimization with a C++ optimization engine. As presented in this paper, \textsc{Submodlib} can be used in a variety of applications like summarization, data subset selection, hyper parameter tuning and efficient training of mdoels. 

\section*{Acknowledgements}

This work is supported in part by the Ekal Fellowship (www.ekal.org), the National Center of Excellence in Technology for Internal Security, IIT Bombay (NCETIS, https://rnd.iitb.ac.in/node/101506) and the IBM AI Horizon Networks. This work is also supported by the National Science Foundation under Grant No. IIS-2106937, a startup grant from UT Dallas, and by a Google and Adobe research award.

\bibliographystyle{plainnat}
\bibliography{main.bib}

\end{document}